\documentclass[letterpaper, 10 pt, conference]{ieeeconf}  

\IEEEoverridecommandlockouts                              

\usepackage{iftex}
\usepackage{mathrsfs}
\usepackage{fontspec}
\usepackage{xeCJK}
\usepackage{booktabs}
\usepackage{multirow}
\usepackage{graphicx} 
\usepackage{placeins}

\defaultfontfeatures{Ligatures=TeX}
\IfFontExistsTF{SimSun}{\setCJKmainfont{SimSun}}{%
        \IfFontExistsTF{Microsoft YaHei}{\setCJKmainfont{Microsoft YaHei}}{%
                \IfFontExistsTF{FandolSong-Regular}{\setCJKmainfont{FandolSong-Regular}}{}%
        }%
}

\usepackage{amsmath} 
\usepackage{amssymb}  
\usepackage{amsfonts}
\usepackage{dblfloatfix}
\usepackage{float}
\usepackage{subcaption}

\usepackage{xcolor}
\usepackage[table]{xcolor}
\usepackage{cite}
\definecolor{cvprblue}{RGB}{0,102,204}  
\usepackage[xetex,colorlinks=true,citecolor=cvprblue,linkcolor=cvprblue,urlcolor=cvprblue]{hyperref}
\definecolor{best}{RGB}{220,235,255}      
\definecolor{second}{RGB}{235,245,255}    

\title{\LARGE \bf
HIMM: Human-Inspired Long-Term Memory Modeling for Embodied Exploration and Question Answering 
}

\author{Ji Li$^{1}$, Bo Wang$^{1}$, Jing Xia$^{1}$, Mingyi Li$^{2}$, Shiyan Hu$^{1,\dagger}$
\thanks{$^{1}$The University of Hong Kong, Hong Kong SAR.
        $^{2}$Beijing Institute of Technology, Beijing, China. $^\dagger$ corresponding author.
        {\tt\small \{jerichojili, bowang, jingxia\}@connect.hku.hk, mingyili@bit.edu.cn, shiyanhu@hku.hk}}%
}

\begin{document}

\maketitle
\thispagestyle{empty}
\pagestyle{empty}

\begin{abstract}
Deploying Multimodal Large Language Models as the brain of embodied agents poses challenging when reasoning long-horizon observations. 
Existing memory-augmented approaches typically compress observations into textual summaries losing fine-grained information, and overlook the inherent differences across memory types.
To address these limitations, we propose HIMM, a memory framework that explicitly models episodic and semantic memory for embodied exploration and question answering. 
Our approach recalls episodic experiences based on semantic similarity and verifies them through visual reasoning on exploration maps, enabling efficient reuse of past observations without strict geometric alignment. 
Meanwhile, we introduce a program-style rule extraction mechanism that transforms experiences into structured, reusable semantic memory, facilitating cross-environment generalization.
Our methods outperforms prior methods on on embodied exploration and question answering benchmarks, yielding a 7.3\% LLM-Match gain and an 11.4\% LLM-Match$\times$SPL gain on A-EQA, as well as +7.7\% success rate and +6.8\% SPL on GOAT-Bench.
Analyses reveal that our episodic memory primarily improves exploration efficiency, while semantic memory strengthens complex reasoning of embodied agents.

\end{abstract}

\section{Introduction}
\label{sec:intro}
In recent years, Multimodal Large Language Models (MLLMs) have advanced rapidly, demonstrating strong reasoning and perception capabilities across a wide range of domains. 
Despite this progress, deploying MLLMs as the core module of embodied agents in real-world environments remains highly challenging. 
A key difficulty lies in enabling agents to efficiently process long-horizon observations while retaining task-relevant information under limited context lengths.
To address this issue, recent works introduce various forms of working memory, allowing agents to retrieve a small subset of past observations instead of conditioning on the entire sensory history~\cite{HOV-SG,DOV-SG,3DMEM}. 
By coupling task-relevant information retrieval with frontier-based exploration strategies, these methods achieve efficient and effective performance in embodied exploration and question answering.

However, most existing approaches remain confined to information within the current episode, limiting their ability to leverage accumulated experience over time. 
In contrast, real-world embodied agents are expected to continuously integrate observations across episodes and exploit long-term knowledge acquired from past interactions. 
Approaches such as~\cite{MemoryEQA, ReEXplore} attempt to address this by summarizing cross-episode information into hierarchical textual descriptions that guide future exploration. 
Meanwhile,~\cite{EnterMindPalace} introduces cross-episode memory into the reasoning loop. 
Nevertheless, its task formulation remains restrictive, typically evaluating an MLLM-based information-collection system within a fixed explored space derived from a previous timeline.
Despite these efforts, existing memory architectures remain fundamentally rigid, and they suffer from two major limitations.
First, this design inevitably discards a substantial amount of spatial and visual cues in raw observations, limiting the agent’s memory capacity.
Second, most methods adopt a single memory representation to model heterogeneous memory systems, which fails to capture the distinct characteristics and functions of different memory types, as extensively studied in cognitive psychology~\cite{SPM}.

\begin{figure*}[t]
    \centering
    \includegraphics[width=\textwidth]{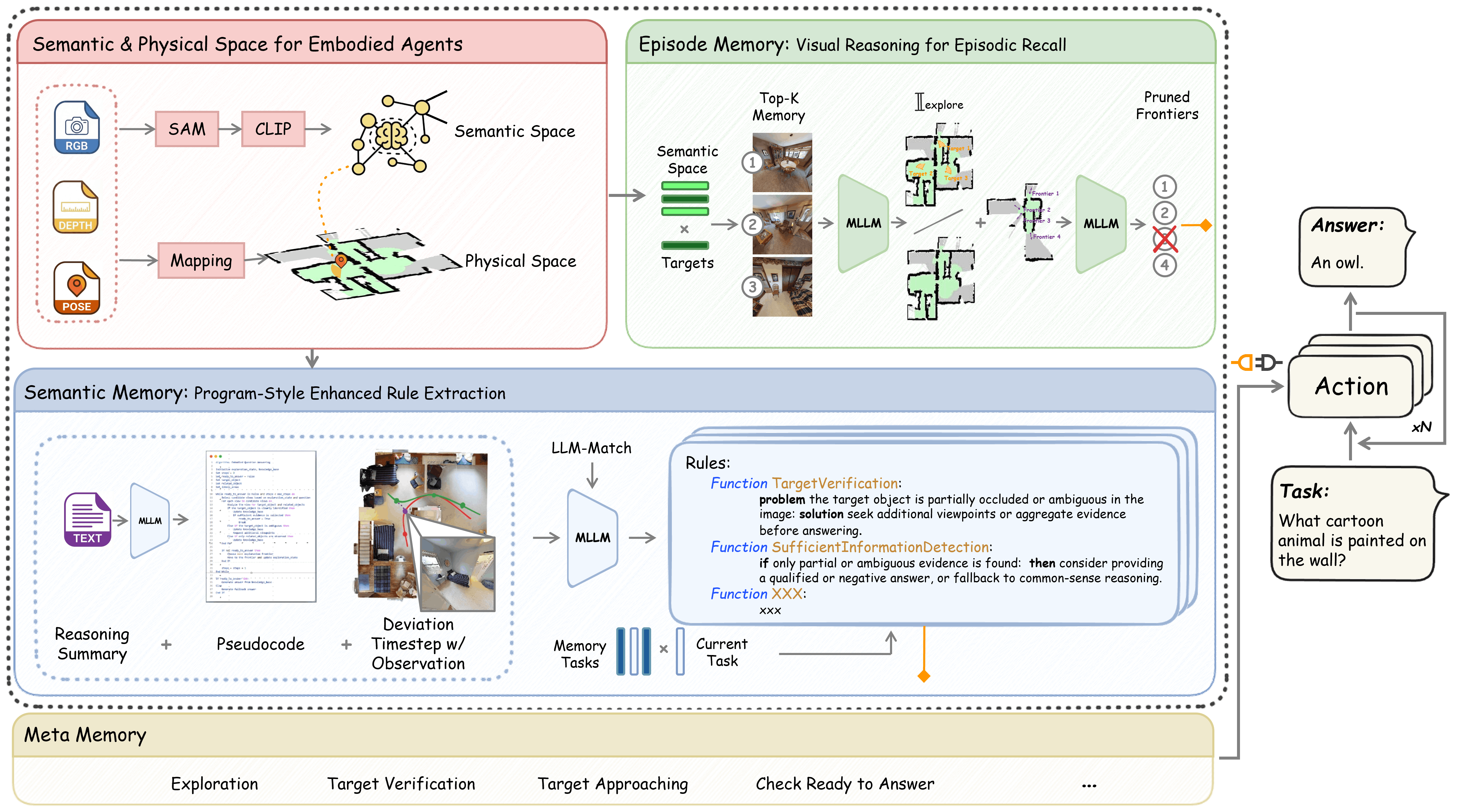}
    \caption{
    Overview of our proposed HIMM.
    Episodic memory and semantic memory guide agent's efficient exploration and accurate reasoning, acrossing sequential cognitive states which governed by meta memory.
    }
    \label{fig:pipeline}
\vspace{-10pt}
\end{figure*}
These limitations become even more pronounced when moving from simulation to realistic dynamic environments, where two additional fundamental challenges arise.
First, fusion maps across time, whether represented as persistent nodes~\cite{DOV-SG} or voxel grids~\cite{Dynam3D}, becomes inherently difficult in the presence of environmental changes and noise. 
Second, information retrieved from past does not necessarily reflect the current state of the environment.
Based on these observations, we argue that embodied agent should neither rely on rigid long-term geometric fusion nor blindly trust past episodic information. 
Instead, it should recall prior experiences in a soft and associative manner while continuously grounding them in active exploration and perception, closely mirroring how humans reason about and interact with dynamic environments.

Motivated by these insights, we propose HIMM, a non-parametric memory framework inspired by human memory systems, which explicitly model episodic and semantic memory for embodied exploration and question answering. 
We propose a retrieval-first and reasoning-assisted paradigm. Retrieved memory snapshot are selectively grounded and verified through visual reasoning, ensuring reliable reuse in long-horizon timeline. 
Complementarily, in contrast to leveraging environment-specific and free-form textual summaries, our method distills structured, high-level, and reusable semantic memory through pseudocode-like decision workflows and decision deviations detection. 
This dual-system design enables stable long-term knowledge accumulation while supporting robust exploration and long-horizon reasoning.
We summarize our contributions as follows:
\begin{itemize}

\item We introduce HIMM, a memory framework that explicitly models episodic and semantic memory for embodied agents, enabling self-evolving without additional training.

\item We propose to ground retrieved episodic memories through visual reasoning over memory snapshots and exploration maps, together with a program-style rule extraction mechanism that enhances complex embodied reasoning.

\item Our method achieves new state-of-the-art results, with 7.3\% and 11.4\% performance gain on A-EQA~\cite{OpenEQA}, and 7.7\%, 6.8\% performance gain on GOAT-Bench~\cite{GOATBench}, demonstrating significantly improved exploration efficiency and remarkable question answering capability.

\end{itemize}

\section{Related Works}

\subsection{MLLM-based Embodied Agents}
Recent proprietary foundation models, such as GPT-5 and Gemini 3, have significantly advanced multi-image understanding by leveraging extended context lengths and large-scale pretraining~\cite{LatentMemory}, and enable more comprehensive perception over long visual sequences, forming a strong backbone for embodied agents \cite{EQA, BeliefMapNav, EfficientNav, MTU3D, GraphEQA,Agent3DZero}.
Beyond vision-language foundation models, imagination driven designs augment observations via generative priors \cite{MindJourney,DreamtoRecall}, while confidence calibration approaches explicitly model when MLLM should stop by quantifying epistemic uncertainty \cite{PruneThenPlan, UntilConfident, BeyondEQA}.

Despite these advances, foundation models still remain computationally expensive, and naively processing uniformly sampled frames is often suboptimal for questions that require selectively identifying informative observations for exploration and reasoning.
To address these challenges, \cite{ConceptGraphs, HOV-SG, DOV-SG, FSR-VLN} introduces CLIP~\cite{CLIP} and SAM-based~\cite{SAM} open-vocabulary scene graphs to structure semantics for MLLM-driven exploration.
While \cite{3DMEM} further incorporates raw images to enrich memory, its key-frame selection remains constrained by detectors and lacks explicit geometric modeling, leading to inefficient long-horizon exploration.
In contrast to existing geometric-semantic graph methods often depend on heavy multi-frame fusion, our approach marks a shift toward decoupling semantic and geometric spaces, enabling retrieval-time interaction that supports adaptive cross-episode recalling.

\subsection{Experience and Memory for Embodied Agents}
Research on memory mechanisms for embodied agents seeks to enable the storage, retention, and retrieval of past experiences, facilitating the transition from purely reactive systems to agents capable of maintaining context and autonomous adaptation.
\cite{ReMember} and \cite{MemoryEQA} adopt descriptive text, temporal and spatial cues to manage continuously growing histories, allowing robots to recall previously visited regions and reduce redundant exploration. 
However, memory built primarily on image and video captioning discard a substantial amount of fine-grained visual information embedded in raw observations. 
\cite{EnterMindPalace} further explores long-term memory under dynamic environments by preserving raw image observations. 
Nevertheless, its task formulation remains restrictive, typically evaluating an information searching system within a fixed explored space, hinders its applicability to realistic exploration scenarios.
More recently, \cite{ReEXplore} focuses on extracting reusable experience from training trajectories. While effective at capturing high-level patterns, the distilled experiences remain tightly coupled to previously environment, limiting their generalization ability and reasoning enhancement.
Motivated by these observations, we draw inspiration from human memory systems and propose to disentangle episodic and semantic memory, recalling episode to improve exploration efficiency, while following high-level decision rules in decision making.

\begin{table*}[t]
\centering
\small
\setlength{\tabcolsep}{3.5pt}
\resizebox{\textwidth}{!}{%
\begin{tabular}{lcccccccccccccccc}
\toprule
\multirow{2}{*}{Method} &
\multicolumn{2}{c}{Object Rec.} &
\multicolumn{2}{c}{Object Loc.} &
\multicolumn{2}{c}{Attribute Rec.} &
\multicolumn{2}{c}{Spatial} &
\multicolumn{2}{c}{Object State} &
\multicolumn{2}{c}{Functional} &
\multicolumn{2}{c}{World Know.} &
\multicolumn{2}{c}{Overall} \\
\cmidrule(lr){2-3} \cmidrule(lr){4-5} \cmidrule(lr){6-7} \cmidrule(lr){8-9}
\cmidrule(lr){10-11} \cmidrule(lr){12-13} \cmidrule(lr){14-15} \cmidrule(lr){16-17}

\multicolumn{17}{l}{\textbf{Socratic LLM-based Exploration w/ Frame Captions}} \\
GPT-4*   & 25.3 & -- & 28.4 & -- & 27.3 & -- & 37.7 & -- & 47.2 & -- & 54.2 & -- & 29.5 & -- & 35.5 \\
GPT-4o   & 22.0 & -- & 25.0 & -- & 27.3 & -- & 40.8 & -- & 50.9 & -- & 61.8 & -- & 38.4 & -- & 35.9 \\

\midrule
\multicolumn{17}{l}{\textbf{Socratic LLM-based Exploration w/ Scene-Graph Captions}} \\
CG Scene-Graph*     & 25.3 & -- & 16.5 & -- & 29.2 & -- & 37.0 & -- & 52.2 & -- & 46.8 & -- & 37.8 & -- & 34.4 \\
SVM Scene-Graph*    & 29.0 & -- & 17.2 & -- & 31.5 & -- & 31.5 & -- & 54.2 & -- & 39.8 & -- & 38.9 & -- & 34.2 \\
LLaVA-1.5*          & 25.0 & -- & 24.0 & -- & 34.1 & -- & 34.4 & -- & 56.9 & -- & 53.5 & -- & 40.6 & -- & 38.1 \\
Multi-Frame*        & 34.0 & -- & 34.3 & -- & 51.5 & -- & 39.5 & -- & 51.9 & -- & 45.6 & -- & 36.6 & -- & 41.8 \\

\midrule
\multicolumn{17}{l}{\textbf{Open-Sourced MLLM-based Exploration}} \\
3D-Mem~\cite{3DMEM} (Qwen2.5-VL) & 25.0 & 13.9 & 23.8 & 7.7 & \cellcolor{best}\textbf{52.6} & 29.3 & 33.3 & 3.9 & 57.5 & \cellcolor{best}\textbf{26.1} & 43.8 & 13.5 & 37.5 & 8.4 & 39.1 & 14.6 \\
ReEXplore~\cite{ReEXplore} (Qwen2.5-VL) 
                    & \cellcolor{best}\textbf{50.6} & \cellcolor{best}\textbf{31.3} & \cellcolor{best}\textbf{29.4} & \cellcolor{best}\textbf{17.8} & 49.1 & \cellcolor{second}31.0 & \cellcolor{best}\textbf{43.1} & \cellcolor{best}\textbf{15.1} & \cellcolor{best}\textbf{66.7} & 17.2 & \cellcolor{second}47.9 & \cellcolor{best}\textbf{27.4} & \cellcolor{second}43.1 & \cellcolor{second}20.9 & \cellcolor{best}\textbf{46.2} & \cellcolor{second}23.0 \\
Ours (Qwen2.5-VL) 
                    & \cellcolor{second}39.0 & \cellcolor{second}29.5 & \cellcolor{second}28.4 & \cellcolor{second}17.3 & \cellcolor{second}51.3 & \cellcolor{best}\textbf{32.4} & \cellcolor{second}37.6 & \cellcolor{second}14.4 & \cellcolor{second}58.3 & \cellcolor{second}24.4 & \cellcolor{best}\textbf{48.5} & \cellcolor{second}24.8 & \cellcolor{best}\textbf{47.2} & \cellcolor{best}\textbf{22.3} & \cellcolor{second}43.1 & \cellcolor{best}\textbf{23.7} \\
\midrule
\multicolumn{17}{l}{\textbf{Commercial MLLM-based Exploration}} \\
Explore-EQA* (GPT-4o) 
                    & 44.0 & 19.6 & 37.1 & 29.6 & 55.3 & 36.0 & 42.1 & 6.6 & 46.3 & 9.2 & 63.2 & 35.7 & \cellcolor{second}45.5 & 22.0 & 46.9 & 23.4 \\
CG + Frontier* (GPT-4o) 
                    & \cellcolor{second}45.0 & \cellcolor{second}42.0 & 32.1 & 25.0 & 50.8 & 35.2 & 32.9 & 18.7 & 68.5 & 38.4 & 58.8 & 42.2 & \cellcolor{second}45.5 & \cellcolor{second}33.5 & 47.2 & 33.3 \\
3D-Mem~\cite{3DMEM} (GPT-4o)     
                    & 35.0 & 18.8 & 50.0 & 37.3 & 64.3 & \cellcolor{best}\textbf{56.3} & 50.0 & 24.7 & 80.0 & 49.3 & 50.0 & 22.1 & 30.0 & 21.4 & 54.4 & 33.3 \\
ReEXplore~\cite{ReEXplore} (GPT-4o)  
                    & 37.5 & 21.9 & \cellcolor{best}\textbf{65.6} & \cellcolor{second}45.8 & \cellcolor{second}67.9 & 46.1 & \cellcolor{best}\textbf{60.0} & \cellcolor{second}25.3 & \cellcolor{best}\textbf{100.0} & \cellcolor{second}53.5 & \cellcolor{second}65.0 & \cellcolor{second}42.4 & 35.0 & 23.6 & \cellcolor{second}58.3 & \cellcolor{second}37.3 \\
Ours (GPT-4o)  
                    & \cellcolor{best}\textbf{62.0} & \cellcolor{best}\textbf{52.7} & \cellcolor{second}63.7 & \cellcolor{best}\textbf{48.6} & \cellcolor{best}\textbf{68.5} & \cellcolor{second}52.5 & \cellcolor{second}52.2 & \cellcolor{best}\textbf{36.8} & \cellcolor{second}91.2 & \cellcolor{best}\textbf{54.1} & \cellcolor{best}\textbf{66.4} & \cellcolor{best}\textbf{45.5} & \cellcolor{best}\textbf{54.5} & \cellcolor{best}\textbf{44.3} & \cellcolor{best}\textbf{65.6} & \cellcolor{best}\textbf{48.7} \\
\midrule
Human Agent*        & 89.7 & -- & 72.8 & -- & 85.4 & -- & 84.8 & -- & 97.8 & -- & 78.9 & -- & 88.5 & -- & 85.1 \\
\bottomrule
\end{tabular}
}
\caption{Performance comparison on A-EQA across different question categories.
For each category, we report the LLM-Match score (M) and the LLM-Match$\times$SPL score.
``CG'' denotes ConceptGraphs. Results marked with * are taken from 3D-Mem~\cite{3DMEM}.}
\label{tab:aeqa_results}
\vspace{-15pt}
\end{table*}

\section{Methods}
Our overall pipeline is illustrated in Fig.~\ref{fig:pipeline}. Given a posed RGB-D observation, our framework constructs semantic and physical space to represent the scene (Section \ref{subsec:SPS}),
Next, episodic memory is recalled by visual reasoning to guide exploration, while program-sytle extracted semantic memory is fed into LLM to improve rational (Section \ref{subsec:Memory}).

\subsection{Semantic and Physical Space for Embodied Agents}
\label{subsec:SPS}

Navigation in embodied environments requires reasoning jointly over a semantic space and a physical space.
Inspired by cognitive science findings suggesting structured interactions between these two spaces~\cite{SPS},
we propose to decouple semantic memory from physical geometry, while enabling their interaction through test-time retrieval.

\begin{figure}[t]
    \centering
    \includegraphics[width=0.8\linewidth]{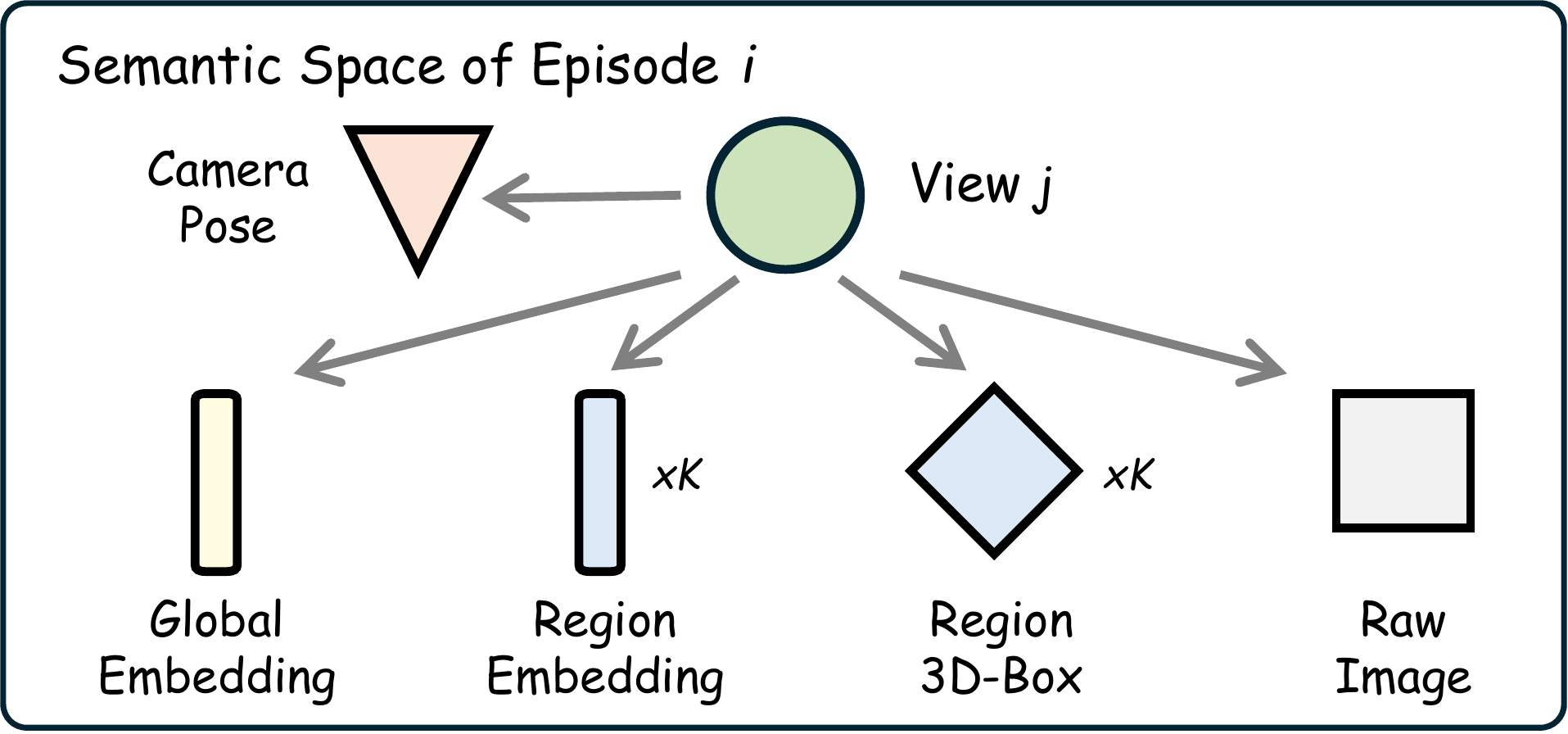}
    \caption{Illustration of our semantic space for embodied agents.}
    \label{fig:semantic_space}
\vspace{-20pt}
\end{figure}

\textbf{Semantic and Physical Space.}
At each observation step, a set of region-level semantic embeddings and a place-level image embedding are extracted using SAM and CLIP to form hierarchical embedding-based scene graph, following HOV-SG \cite{HOV-SG}.
Global embedding, regional embeddings, region 3D-boxes and raw image are stored in a semantic space indexed by camera pose, as depicted in Fig.~\ref{fig:semantic_space}.
To interface with downstream tasks, we leverage a LLM to decompose an instruction into three goal components:
a \emph{target object}, a set of \emph{relative objects}, and a set of \emph{relative areas}.
Each component is embedded and used to query region embeddings in the semantic space.
Retrieved embeddings are ranked in a priority order, with \emph{target object} evidence taking precedence,
followed by \emph{relative objects} and \emph{relative areas}.
While the physical space is represented by a lightweight 2D occupied exploration map constructed from depth observations and camera poses.

\textbf{Retrieval-Time Inter-Space Interaction.}
In constrast to prior approaches such as \cite{HOV-SG}, which tightly couple semantic representations with multi-view geometric fusion,
our semantic memory is updated independently of persistent geometry. 
Specifically, we only link semantic space to the physical space by rendering the camera poses of retrieved observations onto the occupancy grid.
The interaction allowing semantic concepts to be flexibly grounded without continuous online fusion, and preserves task-relevant semantic cues while keeping the physical representation compact and efficient.

\subsection{Human-Inspired Memory for Embodied Exploration}
\label{subsec:Memory}
Embodied agents are expected to learn and adapt from their past experiences.
Episodic and semantic memory play essential roles in human cognition~\cite{SPM}.
This inspires that instead of using a single memory representation to model heterogeneous memory systems, it is important to design different construction and retrieval strategies for each type of memory to better capture their distinct functions.

\textbf{Episodic Memory: Visual Reasoning for Episodic Recall.}
Episodic memory records episode-level experiences in a detailed and temporally grounded manner~\cite{SPM}.
Each episodic memory instance consists of both a semantic space and a physical space, which encode the observed scene from complementary perspectives, as described in Section \ref{subsec:SPS}.
As a fine-grained memory type, episodic memory preserves information at the granularity of navigation steps.

When encountering a new episode, we first retrieve the Top-$K$ most similar past observations based on visual similarity to the current observation.
To verify episodic correspondence, we further assess whether the retrieved observations originate from nearby locations using an MLLM.
After that, embeddings of the current target are used to query the verified episode, and the episode that accumulates the highest number of Top-$K$ matches is selected as prior experience.

Mapping-based approaches~\cite{RoomsFromMotion} explicitly align and merge memories by transformations estimation, which are highly sensitive to alignment failures and are prone to cascading errors during long-horizon changing and noises.
While we humans recall past experiences through associative and semantic similarity rather than rigid metric fusion of memories.
Hence, in contrast to these methods, we propose a gentle episodic recall strategy supported by visual reasoning capability. 

Specifically, given the retrieved Top-$K$ episodic observations, MLLM first determines whether further exploration is required for the current task.
\begin{equation}
\mathbb{I}_{\mathrm{explore}} =
\begin{cases}
1, & \text{if exploration is required}, \\
0, & \text{otherwise}.
\end{cases}
\end{equation}
Let $\mathcal{OCC}^r += (\mathscr{T}^r, \mathcal{F}^r)$ composes the exploration map,
where $\mathscr{T}^r$ represent the retrieved camera poses,
and $\mathcal{F}^r$ denotes the set of frontier points.
For each frontier $f \in \mathcal{F}$, let
$\rho(f)$ indicate whether it can lead to unexplored regions,
and $d(f, \mathscr{T}^r)$ denote its spatial distance to semantically relevant landmarks.
The current and previous exploration maps are jointly provided to a MLLM, outputing pruned frontiers in a visual reasoning manner:
\begin{equation}
\mathcal{F}^\star =
\begin{cases}
\{ f \in \mathcal{F} \mid \rho(f) = 1 \},
& \text{if } \mathbb{I}_{\mathrm{explore}} = 1, \\[6pt]
\{ f \in \mathcal{F} \mid d(f, \mathscr{T}^r) > d_{min} \},
& \text{if } \mathbb{I}_{\mathrm{explore}} = 0 .
\end{cases}
\end{equation}
Here, another retrieval-time interaction between semantic and physical space is expressed through frontiers pruning operation. 
Through this reasoning-centric approach, episodic memory informs exploration decisions without requiring explicit geometric fusion, resulting in a stable exploration strategy.

\textbf{Semantic Memory: Program-Style Enhanced Rule Extraction.}
Semantic memory captures long-term, consolidated knowledge distilled from an agent's past experiences, including reusable knowledge and preferred response patterns~\cite{SPM2}. 
Unlike episodic memories, which rely on fine-grained details and are inherently tied to specific environments, semantic memory aims to extract reusable rules from dilemmas encountered during reasoning execution, with the core objective of capturing solutions that are transferable across environments. 
However, previous experiments~\cite{MemVerse} reveal that, directly summarizing agents' reasoning logs $\mathcal{S}_{\mathrm{n}}$ are often overly entangled with details that fail to reuse and compress essential knowledge.

\begin{figure}[!h]
    \centering
    \begin{subfigure}[b]{\linewidth}
        \centering
        \includegraphics[width=0.95\linewidth]{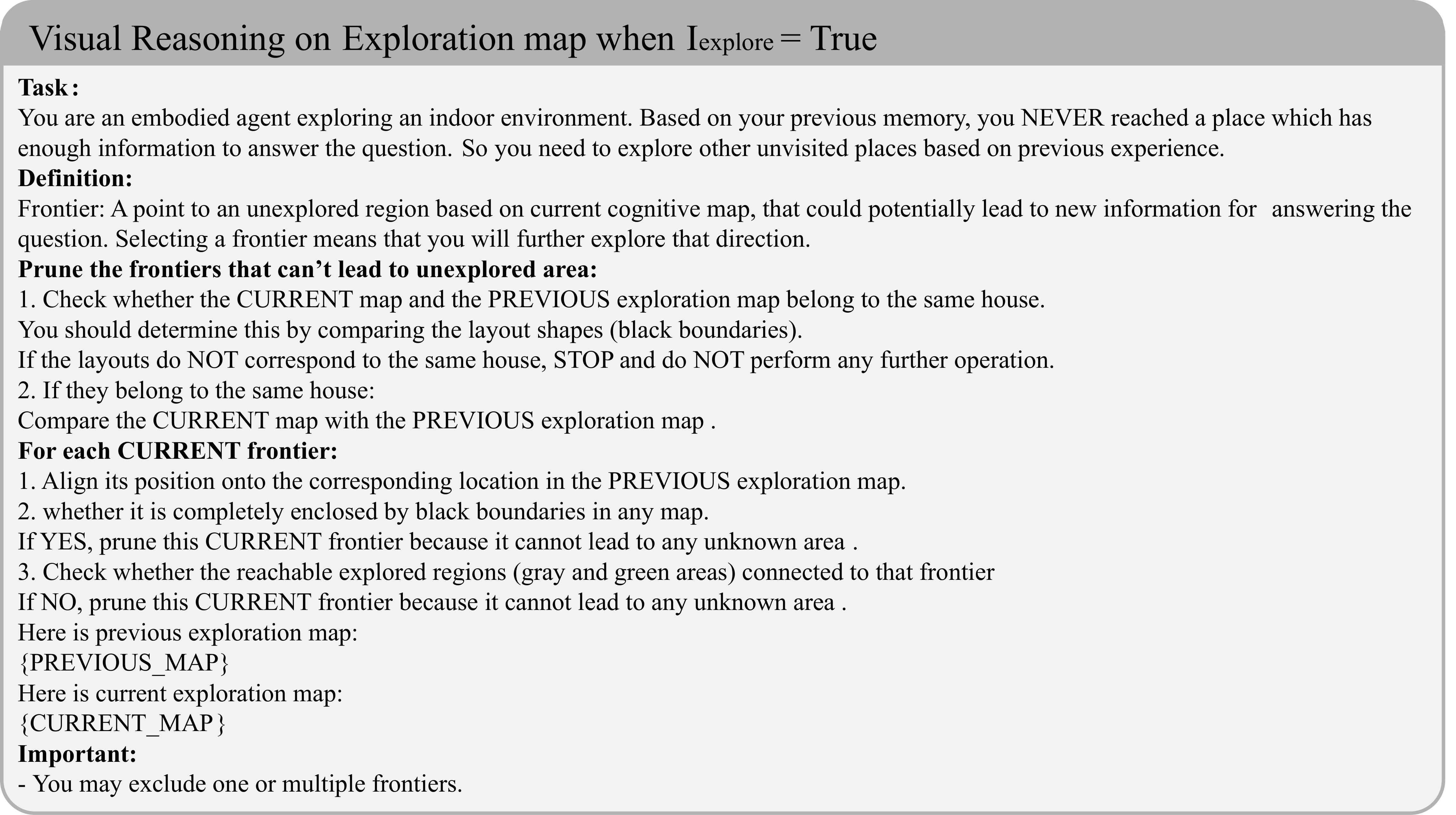}
        \caption{}
        \label{fig:prompt_explore}
    \end{subfigure}
    \\[0pt]
    \begin{subfigure}[b]{\linewidth}
        \centering
        \includegraphics[width=0.95\linewidth]{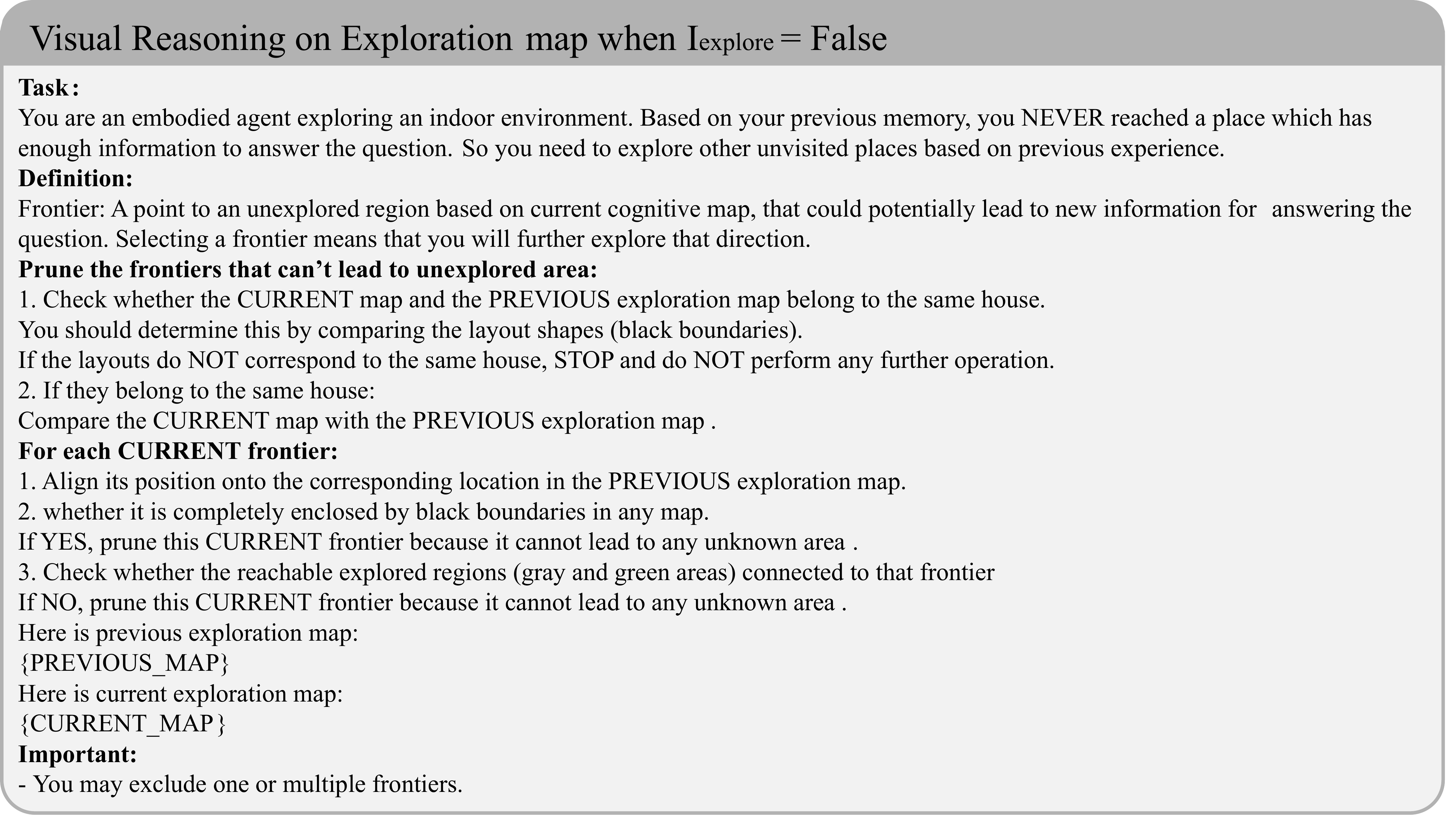}
        \caption{}
        \label{fig:prompt_memory}
    \end{subfigure}
    \caption{
    Prompt templates for visual reasoning on episodic memory recall.
    }
    \label{fig:prompt}
\vspace{-15pt}
\end{figure}

\begin{figure*}[!t]
    \centering
    \includegraphics[width=\textwidth]{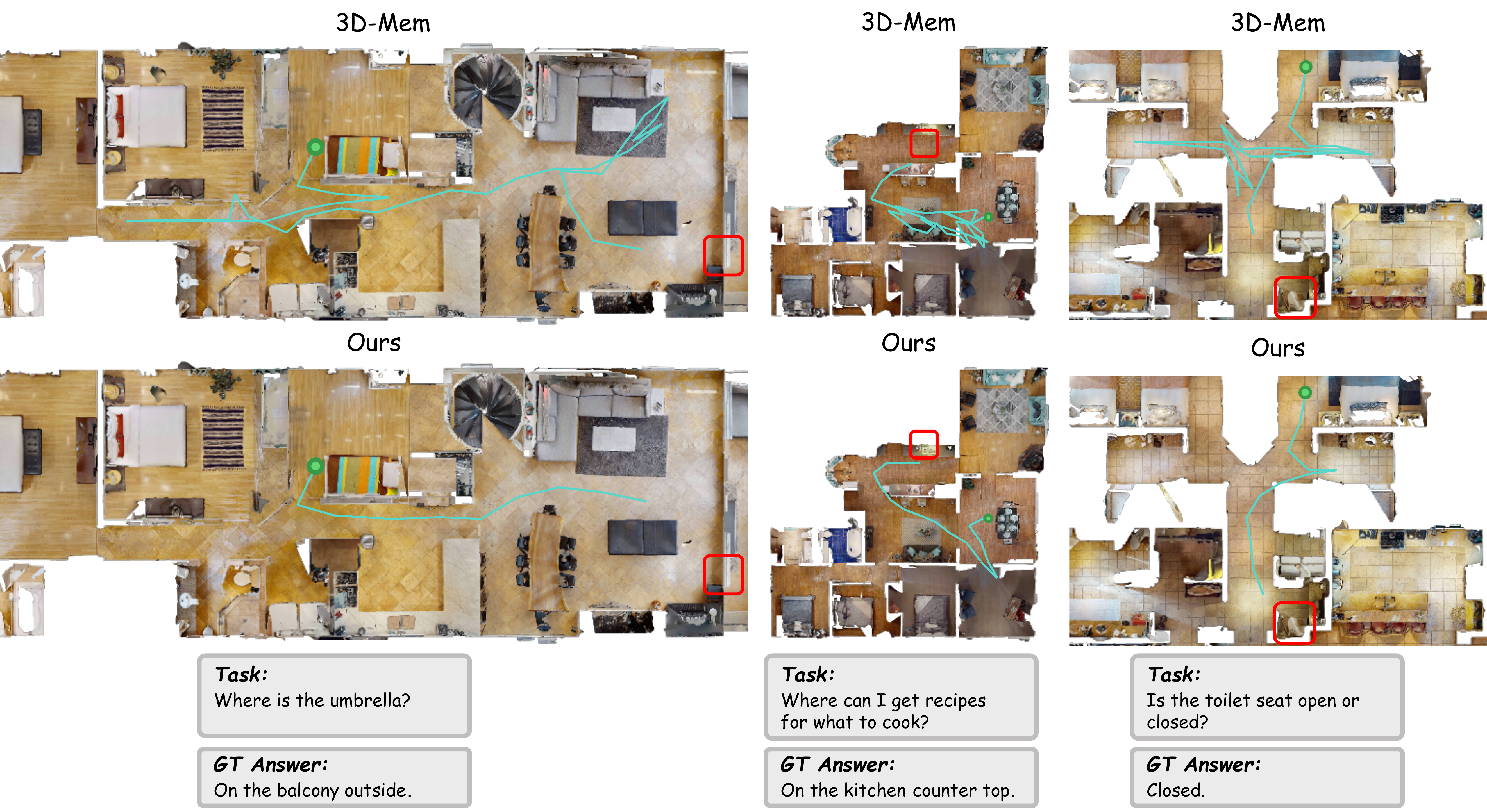}
    \caption{
    Qualitative case study on A-EQA.
    Compared with 3D-Mem, our method demonstrates accurate cross-episode recall and more efficient exploration.
    }
    \label{fig:case-study}
\vspace{-15pt}
\end{figure*}
To address these limitations, we draw inspiration from ~\cite{ArcMemo} and introduce a program-style enhanced rule extraction strategy.
Specifically, we prompt an LLM to analyze the execution pseudocode of a task to prioritize high-level structural information.
The model first identifies \emph{variables} and \emph{functions} that govern the workflows, and then organizes them into a pseudocode $\mathcal{P}$ that explicitly captures control flow and decision logic.
In constract to~\cite{ArcMemo,MemVerse}, where workflows are typically short, embodied agent often involves long-horizon execution with extended decision sequences.
In this situation, LLMs tend to over-annotate functions with learned rules that are weakly correlated with task success, obscuring the truly critical decision points.
To prevent this, we introduce decision deviations detection mechanism on robots' trajectories~\cite{CorrectNav}.
Specifically, given the robot state $\mathbf{M}_t$ at timestep $t$ and the corresponding ground-truth trajectory $x$, we compute:
\begin{equation}
h_t = d(\mathbf{M}_t, x),
\label{eq:deviation}
\end{equation}
where $d(\cdot,\cdot)$ measures the Euclidean distance between the agent's current position and the ground-truth trajectory from training set.
In constract to previous work using predefined threshold, we interactively set threshold $S$ to identify decision deviations at timestep $t$ if
\begin{equation}
h_t > S \;\; \wedge \;\; h_{t-1} \leq S.
\label{eq:threshold}
\end{equation}
Then, we randomly stop the detection at threshold $S_{\mathrm{stop}}$ when the number $K$ of decision deviation is in the range of 3-5 to form set $T=\{t_k\}_{k=1}^{K}$. 
These deviation timesteps signal potential errors or departures from the expected behavior, allowing embodied agent to `debug' the `program'.


During the rule extraction, we provide the MLLM with the ground truth answer $GT$, $\mathcal{S}_{\mathrm{n}}$, $\mathcal{P}$, $t_k$ together with their corresponding visual observations $I_{t_k}$:

\begin{equation}
\begin{aligned}
\mathcal{R}
&=
\mathrm{MLLM}\!\left(
GT,
\mathcal{S}_{\mathrm{n}},
\mathcal{P},
\{(t_k, I_{t_k})\}_{k=1}^{K}
\right), \\
\mathcal{R}
& \leftarrow
\left\{
f_i :
\{\phi_{i,1} \mapsto v_{i,1}, \dots, \phi_{i,n_i} \mapsto v_{i,n_i}\}
\right\}.
\end{aligned}
\end{equation}
Finally, experience semantic knowledge can be abstracted into structured, high-level rules using canonical forms such as \emph{if--then}, \emph{situation--suggestion} and \emph{problem--solution} key-value pairs, where each pair belongs to one specific \emph{variables} or \emph{functions}.

At the test-time, the question-similarity retrieved semantic memory is loaded into the system prompt to constrain decision-making process.
Specifically, MiniLM-L6-v2 variant from Sentence-Transformer~\cite{sentence-transformer} is used to extract embeddings for querying question in semantic memory, and compute cosine similarity to retrieve the top-K.



\textbf{Meta Memory:} 
We leverage the success of cognitive process modeling~\cite{CogNav,EnterMindPalace} and introduce meta memory via cognitive states automatic switching to build the foundation of embodied agent.
Specifically, pre-defined discrete cognitive states include: exploration, target verification, target approaching, check ready to answer.

\begin{figure*}[!h]
    \centering
    \begin{subfigure}[b]{\textwidth}
        \centering
        \includegraphics[width=0.95\textwidth]{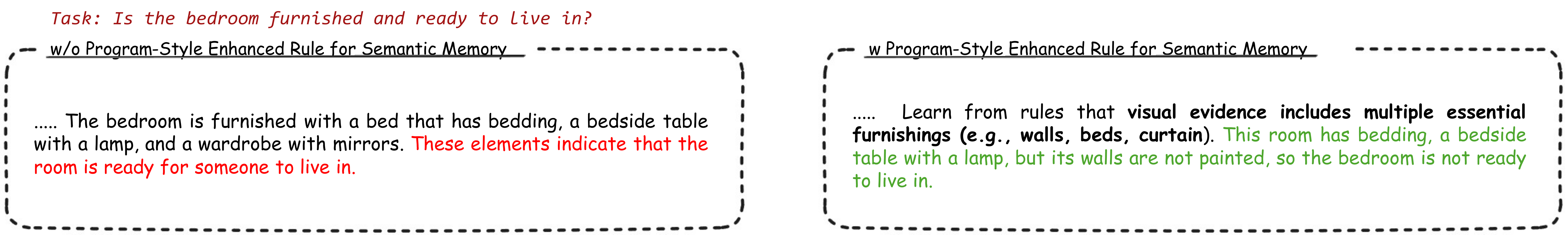}
        \caption{}
        \label{fig:semantic1}
    \end{subfigure}
    \\[-5pt]
    \begin{subfigure}[b]{\textwidth}
        \centering
        \includegraphics[width=0.95\textwidth]{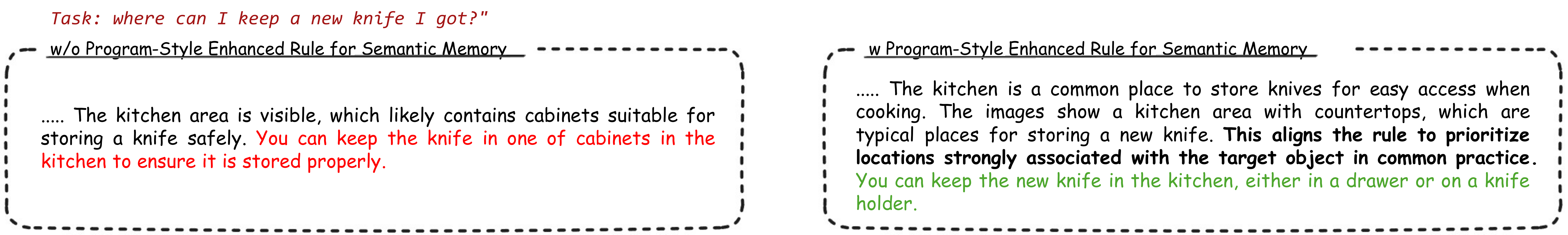}
        \caption{}
        \label{fig:semantic2}
    \end{subfigure}
    \\[-5pt]
    \begin{subfigure}[b]{\textwidth}
        \centering
        \includegraphics[width=0.95\textwidth]{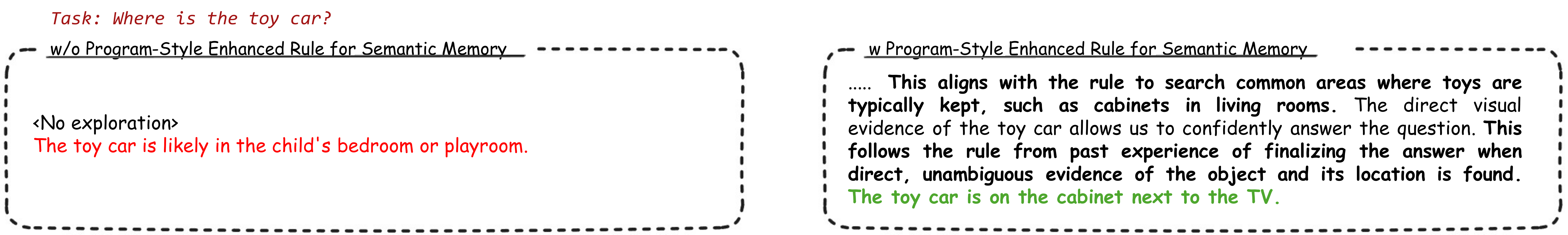}
        \caption{}
        \label{fig:semantic3}
    \end{subfigure}
    \caption{
    Qualitative examples of how program-style enhanced rules for semantic memory impact the reasoning trace for answering questions. We highlight failed reasoning trajectory in red and successful ones in green.
    }
    \label{fig:semantic-case}
\vspace{-15pt}
\end{figure*}

\section{Experiments}
\subsection{Active Embodied Question Answering}
To evaluate the effectiveness of our memory design, we conduct experiments on the A-EQA benchmark~\cite{OpenEQA} which built on habitat~\cite{habitat, hm3d}.
Here, we compare our approach against other MLLM-based baselines on the A-EQA benchmark, assessing both exploration efficiency and performance on answering complex open-ended questions.
In addition, our semantic memory is distilled from 184 episodes in the training split, with a category distribution aligned with that of the test set.


\textbf{Metrics.}
We report \textsc{LLM-Match} and \textsc{LLM-Match}$\times$\textsc{SPL} as evaluation metrics to measure answer quality and exploration efficiency.
Specifically, SPL is computed as the ratio between the shortest path length and the actual trajectory length, normalized to $[0,1]$, where a higher value indicates a more efficient navigation trajectory.
\textsc{LLM-Match} evaluates the semantic correctness of the agent’s answers using GPT-4o as an automated judge.
\textsc{LLM-Match}$\times$\textsc{SPL} further weights the semantic correctness score by navigation efficiency, jointly reflecting both answer quality and exploration effectiveness.
If the agent fails to produce a response, answer is outputted by GPT-4o without visual input, and the SPL term is set to zero.

\begin{table}[h]
\centering
\small
\setlength{\tabcolsep}{6pt}
\resizebox{0.95\columnwidth}{!}{%
\begin{tabular}{lcc}
\toprule
Method & Success Rate (\%)$\uparrow$ & SPL (\%)$\uparrow$ \\
\midrule

\multicolumn{3}{l}{\textbf{Traditional Methods}} \\
Modular GOAT~\cite{GOATBench}*                 & 24.9 & 17.2 \\
Modular CLIP on Wheels~\cite{GOATBench}*   & 16.1 & 10.4 \\
SenseAct-NN Skill Chain~\cite{GOATBench}*      & 29.5 & 11.3 \\
SenseAct-NN Monolithic~\cite{GOATBench}*       & 12.3 & 6.8 \\

\midrule
\multicolumn{3}{l}{\textbf{MLLM-based Exploration}} \\
3D-Mem~\cite{3DMEM}$^\dagger$ (Qwen2.5-VL-7B-Instruct) & 49.4 & 20.7 \\
ReEXplore~\cite{ReEXplore}$^\dagger$ (Qwen2.5-VL-7B-Instruct)          & \cellcolor{second}53.2 & \cellcolor{best}\textbf{32.6} \\
Ours$^\dagger$ (Qwen2.5-VL-7B-Instruct)          & \cellcolor{best}\textbf{53.9} & \cellcolor{second}31.7 \\
Explore-EQA~\cite{UntilConfident}$^\dagger$ (GPT-4o)     & 55.0 & 37.9 \\
CG w/ Frontier Snapshots~\cite{3DMEM}$^\dagger$ (GPT-4o) & 61.5 & 45.3  \\
3D-Mem w/o Memory~\cite{3DMEM}$^\dagger$ (GPT-4o)    & 58.6 & 38.5 \\
3D-Mem~\cite{3DMEM}$^\dagger$ (GPT-4o)               & \cellcolor{second}68.9 & \cellcolor{second}48.9 \\
ReEXplore~\cite{ReEXplore}$^\dagger$ (GPT-4o)                         & 59.8 & 42.5 \\
Ours$^\dagger$ (GPT-4o)                         & \cellcolor{best}\textbf{72.8} & \cellcolor{best}\textbf{56.1} \\

\bottomrule
\end{tabular}
}
\caption{Results on GOAT-Bench under the \textit{Val Unseen} split.
``CG'' denotes ConceptGraphs~\cite{ConceptGraphs}.
Methods marked with * are reported from GOAT-Bench, while those marked with $\dagger$ use MLLM-based exploration.}
\label{tab:goatbench_results}
\vspace{-10pt}
\end{table}

\textbf{Analysis.}
Table~\ref{tab:aeqa_results} shows that our method surpasses all baselines with the GPT-4o backbone. 
Specifically, the overall LLM-Match improves from 58.3\% to 65.6\%, and LLM-Match×SPL increases substantially from 37.3\% to 48.7\%, indicating both higher answer accuracy and improved exploration efficiency.

\begin{table*}[!t]
\centering
\scriptsize
\setlength{\tabcolsep}{3pt}
\renewcommand{\arraystretch}{0.9}
\resizebox{0.85\textwidth}{!}{%
\begin{tabular}{lcc|cc}
\toprule
Method &
Episodic Memory &
Semantic Memory &
LLM-Match &
LLM-Match$\times$SPL \\
\midrule
\textbf{Ours (Qwen2.5-VL-7B-Instruct)}
& \checkmark & \checkmark
& 43.1 & 23.7 \\
& \checkmark & 
& 41.6 & 23.2 \\
&  & \checkmark
& 39.1 & 17.5 \\
\midrule
\textbf{Ours (GPT-4o)}
& \checkmark & \checkmark
& 65.6 & 48.7 \\
& \checkmark & 
& 62.3 & 47.2 \\
&  &  \checkmark
& 60.9 & 41.1 \\
\bottomrule
\end{tabular}
}
\caption{Ablation studies on our method by selectively disabling episodic or semantic memory.
Results are reported with two backbone MLLMs.}
\label{tab:ablation_two_modules}
\vspace{-15pt}
\end{table*}

Notably, significant gains are observed across most task categories: Object recognition improves from 45.0\% to 62.0\%, functional reasoning from 65.0\% to 66.4\%, and world knowledge from 35.0\% to 54.5\%.
On the one hand, our method achieves higher LLM-Match×SPL across all categories. 
As further illustrated in Fig.~\ref{fig:case-study}, episodic memory recall guided by visual reasoning effectively steers exploration toward goal-relevant regions, resulting in more purposeful navigation.
On the other hand, the pronounced improvements in world knowledge and functional recall highlight our semantic memory supporting the reasoning-intensive and cognitively complex tasks.
Fig.~\ref{fig:semantic-case} illustrates how semantic memory constrains the agent’s reasoning during question answering.
Task-relevant semantic memory encapsulates high-level prior experience, reducing the arbitrariness of MLLM reasoning and constraining the agent to adopt more appropriate strategies, thereby improving answering performance.

When switching to the Qwen-2.5-VL-7B-Instruct backbone, our method maintains consistent performance gains. 
It achieves the second-best LLM-Match and the best LLM-Match×SPL, indicating a favorable balance between answer accuracy and exploration efficiency across different MLLM backbones.
In contrast to ReExplore~\cite{ReEXplore}, which employs a stronger MLLM backbone but exhibits noticeable performance degradation in object recognition and world knowledge, our framework continues to scale positively with increasing MLLM capability. 
This suggests that our design more effectively leverages enhanced MLLM reasoning capacity, rather than being bottlenecked by other components in system.

\subsection{Life-Long Visual Language Navigation}
GOAT-Bench~\cite{GOATBench} is a lifelong visual navigation benchmark designed around a multi-target navigation task.
In this setting, an agent is required to navigate sequentially to multiple target objects within previously unseen environments.
Targets are specified in diverse modalities, including object category names (e.g., \emph{refrigerator}), natural language descriptions (e.g., \emph{toy next to the dresser}), or goal images.
Following prior baseline methods~\cite{ReEXplore,3DMEM}, we evaluate our approach on a $1/10$-scale subset of the val-unseen split, which contains one episode per scene across 36 scenes, resulting in a total of 278 navigation subtasks.
In addition, our semantic memory is constructed on the 36 training scenes.

\textbf{Metrics.}
Success Rate and Success weighted by Path Length (SPL) is adpoted as evaluation metrics. 
An episode is deemed successful when the agent terminates within 1 m of the navigation target. 
Similar to OpenEQA, SPL measures efficiency by weighting successful episodes according to the length of the navigation path.

\textbf{Analysis.}
As shown in Table~\ref{tab:goatbench_results}, our method outperforms most baseline methods, demonstrating its ability to effectively leverage knowledge accumulated from both episode memory and semantic memory, as evidenced by consistently higher Success Rate and SPL scores.
Compared with 3D-Mem under the GPT-4o backbone, our method achieves a notable improvement in life-long navigation. The Success Rate increases from 68.9\% to 72.8\%, and SPL rises from 48.9\% to 56.1\%, indicating more accurate and efficient goal-directed exploration.
Under the Qwen backbone, our method slightly underperforms ReEXplore in SPL metric. 
We attribute this primarily to the relatively weaker visual reasoning capability of Qwen backbone when operating on semantic-labeled exploration maps, which limits the correctness of frontier pruning.




\subsection{Ablation Study}
Table~\ref{tab:ablation_two_modules} presents ablation results on the A-EQA benchmark.
When episodic memory is removed, we observe a clear degradation in exploration efficiency. 
Specifically, LLM-Match drops from 41.3\% to 39.1\% under the Qwen backbone, and from 65.6\% to 60.9\% under the GPT-4o backbone. 
This performance decrease indicates that our episodic memory provides effective guidance for exploration by enabling the agent to reuse relevant prior experiences.
In contrast, removing semantic memory results in a more decline in answer accuracy than corresponding reduction in SPL. 
This highlights the critical role of semantic memory in preserving transferable knowledge and supporting embodied reasoning during question answering.
Overall, these ablation results demonstrate that the two memory modules offer complementary benefits: episodic memory primarily improves exploration efficiency, while semantic memory contributes more substantially to accuracy of reasoning-intensive question answering.



\section{Conclusion}
We presented HIMM, a novel memory framework for MLLM-based embodied agents that explicitly model episodic and semantic memory to address long-horizon exploration and reasoning.
By constructing a visual reasoning paradigm, our approach enables agents to recall and verify past episodic experiences without relying on rigid geometric fusion, thereby supporting robust cross-episode reuse of observations and improving exploration efficiency.
In addition, resuable and structed rules as semantic memory distilled from experiences tackle complex reasoning in embodied tasks. Comprehensive evaluation on A-EQA and GOAT-Bench demonstrates the effectiveness of our memory modeling in improving both exploration efficiency and answering accuracy.





\bibliographystyle{IEEEtran} 
\bibliography{IEEEexample}

\end{document}